\documentclass[conference, nonumber]{IEEEtran}
\IEEEoverridecommandlockouts
\usepackage[numbers, sort&compress]{natbib}
\usepackage{hyperref}
\usepackage[acronym]{glossaries}
\usepackage{amsmath,amssymb,amsfonts}
\usepackage{algorithmic}
\usepackage{tabularx}
\usepackage{enumerate}
\usepackage{url}
\usepackage{graphicx}
\usepackage{comment}
\usepackage{caption}
\usepackage{subcaption}
\usepackage{siunitx}
\usepackage{multirow}
\usepackage{placeins}
\usepackage{textcomp}
\usepackage{xcolor}
\usepackage{cleveref}
\crefname{figure}{fig.}{figs.}
\Crefname{figure}{Fig.}{Figs.}
\crefname{table}{tab.}{tabs.}
\Crefname{table}{Tab.}{Tabs.}
\usepackage[a4paper, total={184mm,239mm}]{geometry}
\def\BibTeX{{\rm B\kern-.05em{\sc i\kern-.025em b}\kern-.08em
    T\kern-.1667em\lower.7ex\hbox{E}\kern-.125emX}}
\usepackage{listings}
\newcommand{\todo}[2][TODO: ]{\textcolor{orange}{#1 #2}}

\newacronym{pf}{PF}{Particle Filter}
\newacronym{ekf}{EKF}{Extended Kalman Filter}
\newacronym{mcl}{MCL}{Monte-Carlo Localization}
\newacronym{amcl}{AMCL}{Adaptive Monte-Carlo Localization}
\newacronym{slam}{SLAM}{Simultaneous Localization and Mapping}
\newacronym{pgo}{PGO}{Pose Graph Optimization}
\newacronym{mit}{MIT}{the Massachusetts Institute of Technology}
\newacronym{ros}{ROS}{Robot Operating System}
\newacronym{sota}{SotA}{State of the Art}
\newacronym{lidar}{LiDAR}{Light Detection and Ranging}
\newacronym{imu}{IMU}{Inertial Measurement Unit}
\newacronym{icp}{ICP}{Iterative Closest Point}
\newacronym{ndt}{NDT}{Normal Distributions Transform}
\newacronym{tum}{TUM}{Technical University of Munich}
\newacronym{gpu}{GPU}{Graphics Processing Unit}
\newacronym{cpu}{CPU}{Central Processing Unit}
\newacronym{obc}{OBC}{On Board Computer}
\newacronym{lut}{LuT}{Lookup Table}

\usepackage{eso-pic}
\usepackage{eso-pic}

\newcommand\AtPageUpperMyleft[1]{\AtPageUpperLeft{%
\put(\LenToUnit{1cm},\LenToUnit{-2cm}){#1}%
}}%

\AddToShipoutPictureBG*{%
  \AtPageUpperMyleft{\parbox[b][2cm][c]{\paperwidth}{%
    \centering
    \fontsize{12}{14}\selectfont
    \color{gray!50}
    This paper has been accepted for publication at the\\
    IEEE Design, Automation and Test in Europe Conference, Valencia 2024. \copyright{}IEEE
  }}%
}

\AddToShipoutPictureBG*{
  \AtPageLowerLeft{%
    \raisebox{25pt}{\makebox[\paperwidth]{\begin{minipage}{21cm}\centering
    \fontsize{10}{12}\selectfont
      \textcolor{gray!50}{ \copyright 2024 IEEE.  Personal use of this material is permitted.  Permission from IEEE must be obtained for all other uses, in any current or future media, including reprinting/republishing this material for advertising or promotional purposes, creating new collective works, for resale or redistribution to servers or lists, or reuse of any copyrighted component of this work in other works.
      }
    \end{minipage}}}%
  }
}

\begin{document}
\pagestyle{plain}

\title{Robustness Evaluation of Localization Techniques for Autonomous Racing
{\footnotesize 
}
}

\author{\IEEEauthorblockN{Tian Yi Lim$^{*}$\thanks{$^{*}$Contributed Equally}, Edoardo Ghignone$^{*}$, Nicolas Baumann$^{*}$, Michele Magno}
\IEEEauthorblockA{\textit{Center for Project-Based Learning, ETH Zürich} \\
\{tialim, eghignone, nibauman, magnom\}@ethz.ch} 
}

\maketitle

\begin{abstract}
This work introduces \emph{SynPF}, an MCL-based algorithm tailored for high-speed racing environments. Benchmarked against \emph{Cartographer}, a state-of-the-art pose-graph SLAM algorithm, \emph{SynPF} leverages synergies from previous particle-filtering methods and synthesizes them for the high-performance racing domain. Our extensive in-field evaluations reveal that while \emph{Cartographer} excels under nominal conditions, it struggles when subjected to wheel-slip—a common phenomenon in a racing scenario due to varying grip levels and aggressive driving behaviour. Conversely, \emph{SynPF} demonstrates robustness in these challenging conditions and a low-latency computation time of \SI{1.25}{\milli \second} on on-board computers without a GPU. Using the F1TENTH platform, a 1:10 scaled autonomous racing vehicle, this work not only highlights the vulnerabilities of existing algorithms in high-speed scenarios, tested up until \SI{7.6}{\metre \per \second}, but also emphasizes the potential of \emph{SynPF} as a viable alternative, especially in deteriorating odometry conditions. Code is available here: \url{https://github.com/ForzaETH/particle_filter}.

\end{abstract}

\begin{IEEEkeywords}
Autonomous Racing, Localization, Particle Filter, Monte Carlo Localization, SLAM, Robustness
\end{IEEEkeywords}


\section{Introduction}

Localization approaches for autonomous racing, such as pose-graph based Simultaneous Localization and Mapping (SLAM) \cite{Cartographer} and Monte-Carlo Localization (MCL)-based (also called Particle Filtering, or PF) methods \cite{probabilisticrobotics, MIT_PF, ros-high-speed} depend on both \emph{exteroceptive} and \emph{proprioceptive} inputs. For example, LiDAR sensors offer range measurements for \emph{exteroceptive} sensing, enabling the robot to perceive its environment. In contrast, \emph{proprioceptive} measurements provide insight into the robot's internal states, processing signals from IMUs and wheel-odometry. Consequently, SLAM algorithms can map environments while localizing the robot. On the other hand, MCL-based techniques, relying on both sensing modalities and a pre-existing map, solely determine the robot localization using MCL \cite{MIT_PF, ros-high-speed}.

This paper leverages the F1TENTH autonomous racing platform to compare the performance of pose-graph optimization-based SLAM algorithms against MCL-based algorithms, considering the wheel-odometry signal quality to evaluate the performance in terms of robustness and latency. The robustness measurements can be qualitatively evaluated through scan alignment, the latter latency concern is quantitatively measurable, ultimately leading to a scan matching computation time of \SI{1.25}{\milli \second}. Specifically, we compare the state-of-the-art (SOTA) pose-graph based SLAM algorithm \emph{Cartographer} \cite{Cartographer} with the newly proposed \emph{SynPF}. The latter builds, combines, and synergizes upon prior MCL-based techniques \cite{MIT_PF, ros-high-speed}. 

\section{SynPF Implementation Details}\label{sec:synpf_details}

\begin{figure}[h]
    \centering
    \includegraphics[trim=0 0.5 0 0.55cm, clip, width=0.85\columnwidth]{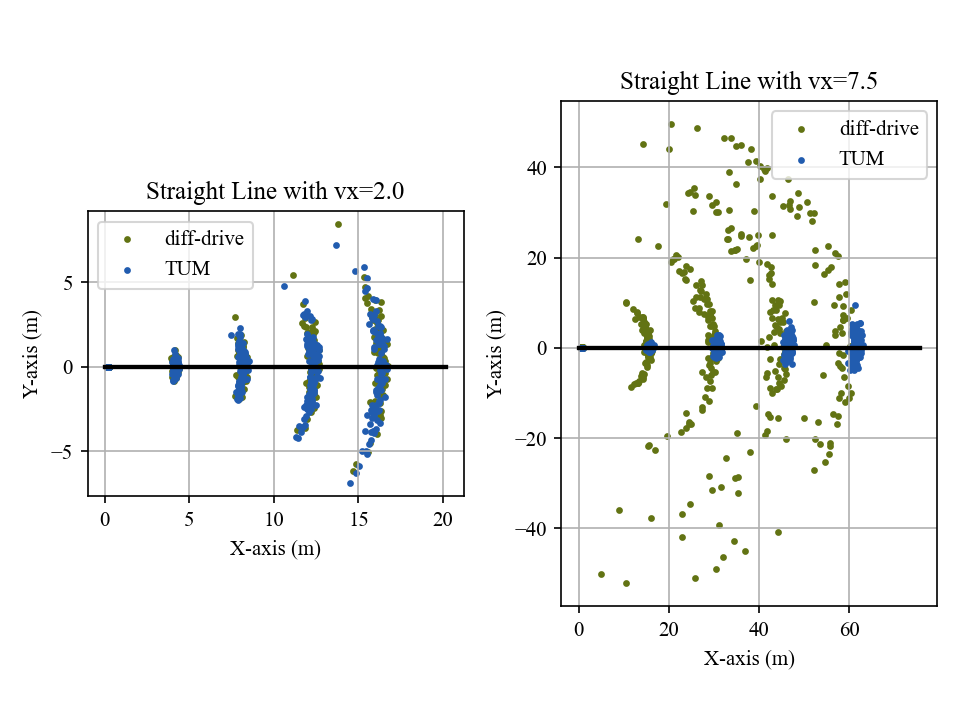}
    \vspace{-2mm}
    \caption{Comparison of poses generated by diff-drive \cite{probabilisticrobotics} and TUM motion models \cite{ros-high-speed}. The left figure shows that at slow speeds, the two models are very similar. The right figure highlights how the TUM motion model further accounts for the reduced steering capacity at higher speeds.}
    \label{fig:mm_selection}
\end{figure}

A commonly used motion model for MCL-based methods is the differential drive (diff-drive) model \cite{probabilisticrobotics} which approximates the steering geometry of a car. This approximation, however, generates unrealistically high angular uncertainties at high speed, resulting in particles being in infeasible positions, reducing \textit{particle efficiency}. 
An improvement used in \emph{SynPF} is described in \cite{ros-high-speed} (hereafter referred to as the TUM PF) which improves this approximation at high longitudinal velocities. This model considers the reduced lateral action space at high velocities, leading to a more realistic pose distribution for high-speed racing. This is motivated by the observation that at high velocities, for instance on straight sections of a racetrack, the car’s angular uncertainty is low, as the steering input cannot be large.
The improvement in the diff-drive model is further illustrated in \cref{fig:mm_selection}.


The TUM PF \cite{ros-high-speed} also proposes a \emph{boxed} LiDAR layout. With a boxed layout, scanlines are chosen such that their intersections with a corridor of configurable aspect ratio are uniformly spaced. This is motivated by the observation that race tracks are corridor-like environments. By spacing the scanlines in this way, the boxed scanlines point further ahead down the racetrack, giving more information on the geometry of the racetrack further ahead. This results in more information with a constant number of scanlines.

In addition, a large proportion of computation effort in MCL methods is in evaluating the expected sensor range at a given pose. This is accelerated by using the \texttt{rangelibc} \cite{MIT_PF} library, which offers two modes of interest. Firstly, it allows the use of an available GPU to parallelize a ray-casting operation over all selected scanlines. This vastly speeds up the calculation of the expected ranges. Secondly, it offers a lookup table (LUT) option to pre-calculate all expected ranges for a discretized set of poses in the map. A 3D array is constructed, with entries for each possible $x,y$ position of the LiDAR and orientation $\theta$ of the scanline. This results in constant-time query speed at the expense of memory usage. Utilizing one of the two methods greatly increases the rate at which the sensor model can be evaluated.

\section{Experimental Results}

\begin{figure}[h]
    \centering
    \begin{subfigure}{.24\textwidth}
        \includegraphics[trim={0cm 0 0cm 25cm}, clip, width=0.99\textwidth]{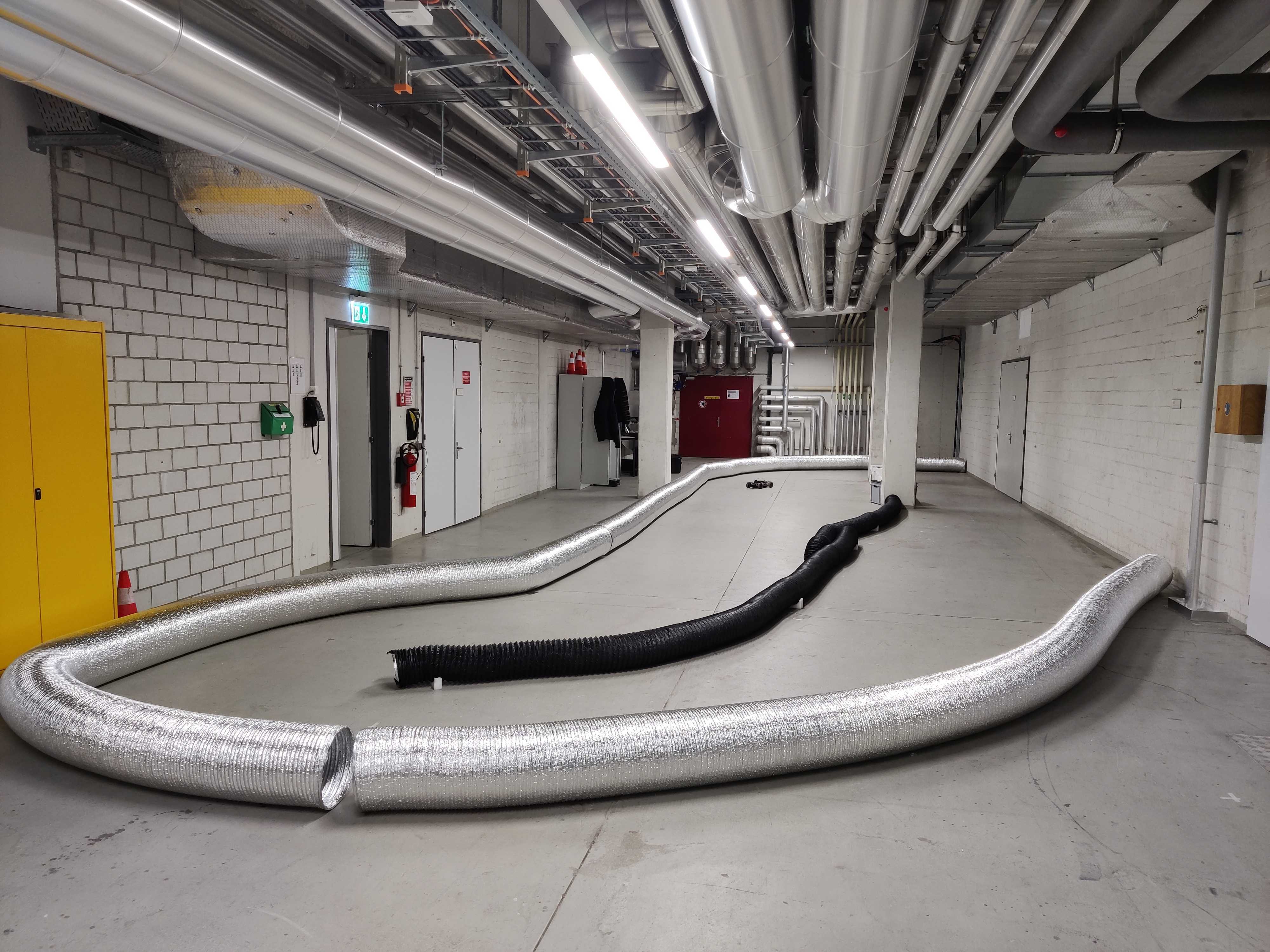}
    \end{subfigure}
    \begin{subfigure}{.24\textwidth}
        \includegraphics[trim={0cm 0 0 0}, clip, width=0.99\textwidth]{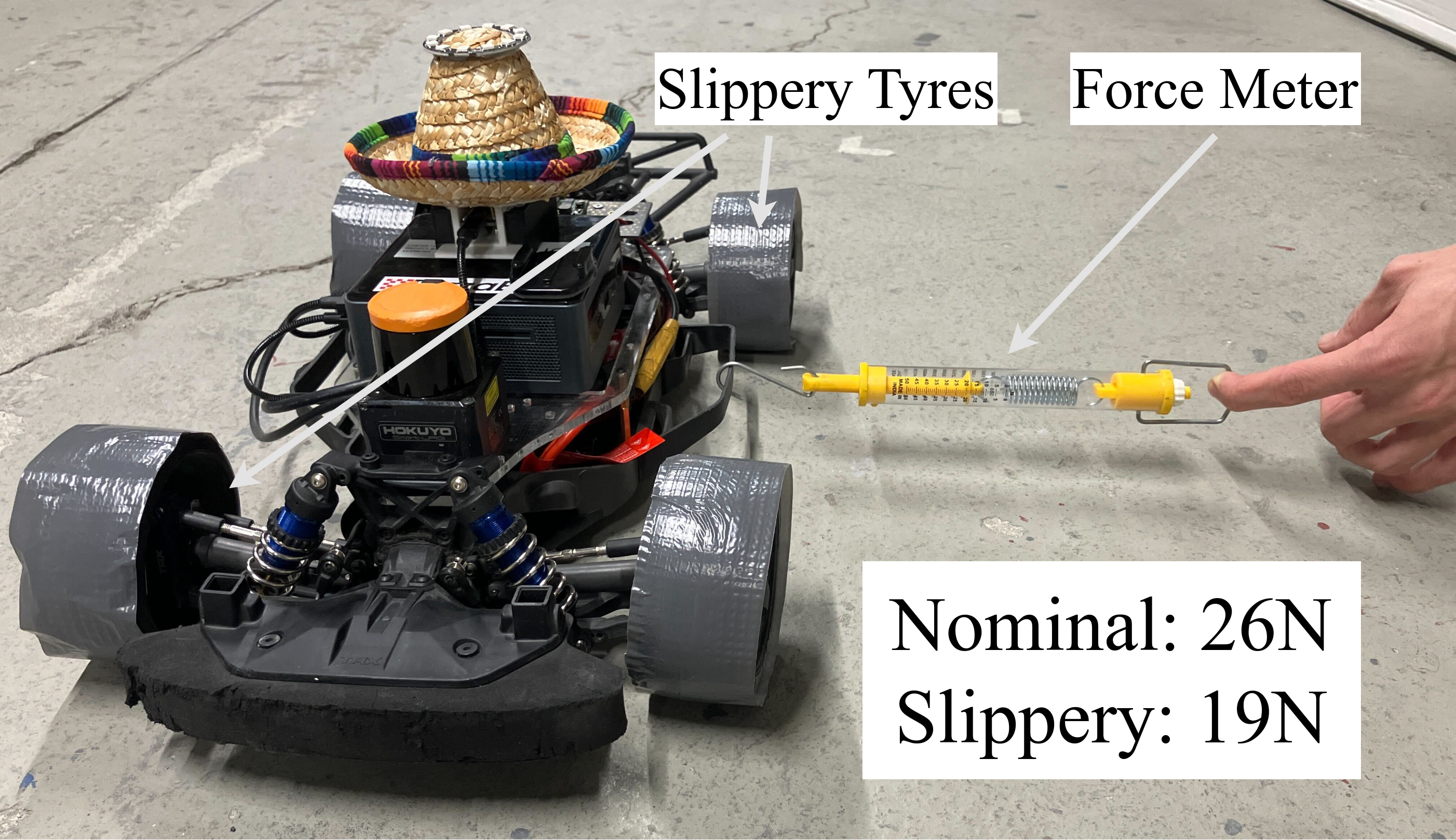}
    \end{subfigure}
    \caption{Test track used for quantitative localization accuracy evaluations. Left: Utilized test track, featuring grippy and high-quality odometry. Right: Racecar with taped and slippery tires, featuring low-quality odometry on the same test track.}
    \label{fig:quant_tracks}
\end{figure}

\Cref{fig:quant_tracks} depicts the test track that was used to quantitatively evaluate the performance of either \emph{Cartographer} or \emph{SynPF}, with respect to the quality of the odometry input. The grip level was measured by pulling the test car laterally along the center of mass.
As the test track features a surface with high friction (\SI{26}{\newton} nominal conditions), the car's tires were altered such that they feature significantly reduced static friction (\SI{19}{\newton} slippery conditions), by applying tape to the tires.
This allows for mimicking the odometry quality degradation similar to a slippery floor. To isolate the odometry degradation effect, 10 laps were completed at the same speed scaling in both settings, where the lap time, scan alignment, and lateral deviation serve as proxy measurements for localization accuracy. The processor unit on the cars used was an Intel NUC on-board computer with an \texttt{i5-10210U} processor, without a dedicated GPU. Therefore, the LUT option in \texttt{rangelibc} \cite{MIT_PF} was utilized for this experiment.

\begin{table}[!h]
\centering
\caption{Lap time and computation results on the test track, featuring slippery (low-quality: LQ) and grippy (high-quality: HQ) odometry input. The average lateral error is with respect to the ideal race line. The scan alignment score is computed by the average percentage of overlapping scans and the track boundary. The compute metric refers to \texttt{htop} percentage of CPU core utilization.}
\label{tab:combined_quant_results}
\setlength{\tabcolsep}{1.5pt} 
\renewcommand{\arraystretch}{1.5} 

\begin{tabular}{|c|c|c|c|c|c|c|c|c|c|}
\hline
\multirow{2}{*}{\textbf{Method}} & \multirow{2}{*}{\textbf{Odom}} & \multicolumn{2}{c|}{\textbf{Lap Time [s]}} & \multicolumn{2}{c|}{\textbf{Error [cm]}} & \multirow{2}{*}{\parbox[c]{1.475cm}{\centering \textbf{Scan} \\ \textbf{Align [\%] $\uparrow$}}}
 & \multirow{2}{*}{\parbox[c]{1.2cm}{\centering \textbf{Load} \\ \textbf{avg $\downarrow$}}} \\
\cline{3-6}
& & $\mu$ $\downarrow$ & $\sigma$ $\downarrow$ & $\mu$ $\downarrow$ & $\sigma$ $\downarrow$ & & \\
\hline
\multirow{2}{*}{\emph{Cartographer}} & HQ & \textbf{9.167} & \textbf{0.097} & \textbf{6.864} & \textbf{0.264} & 69.357 & \multirow{2}{*}{4.2} \\
\cline{2-7}
& LQ & 9.428 & 0.126 & 11.432 & \textbf{1.134} & 61.710 &\\
\hline
\multirow{2}{*}{\emph{SynPF}} & HQ & 9.184 & 0.153 & 8.223 & 0.406 &\textbf{80.603} & \multirow{2}{*}{\textbf{2.17}} \\
\cline{2-7}
& LQ & \textbf{9.280} & \textbf{0.093} & \textbf{7.686} & 1.179 & \textbf{79.924} & \\
\hline
\end{tabular}
\end{table}

\Cref{tab:combined_quant_results} holds the resulting accuracy proxy measurements for \emph{Cartographer} and \emph{SynPF} under the influence of varying odometry quality. \emph{SynPF} shows an approximately \SI{0.15}{\second} faster lap time in the low-quality odometry scenario and a significantly lower lateral deviation of \SI{7.68}{\centi\metre} as opposed to \emph{Cartographer}'s \SI{11.43}{\centi\metre} error. 
On the other hand, when performing the same experiment on the test track under favourable grippy (high-quality) odometry conditions, \cref{tab:combined_quant_results} showcases that the high-quality odometry signal allows \emph{Cartographer} to perform a roughly \SI{0.02}{\second} faster average lap time than \emph{SynPF} with a lower lateral error of \SI{6.86}{\centi\metre}, as opposed to \emph{SynPF}'s \SI{8.22}{\centi\metre} error. 
Interestingly, for \emph{Cartographer} the lateral error increases by 67\% (\SI{6.86}{\centi\metre} to \SI{11.43}{\centi\metre}) when going from high to low-quality odometry, while for \emph{SynPF} the change is merely 6.9\% from \SI{8.22}{\centi\metre} to \SI{7.68}{\centi\metre}.

\FloatBarrier
\section{Conclusion}
This work introduced \emph{SynPF}, an MCL-based localization algorithm for autonomous racing that synergizes the efforts made in previous MCL-based approaches \cite{MIT_PF, ros-high-speed}. 
The novel algorithm integrates a motion and sensor model that more accurately describes the high-speed racing scenario. \emph{SynPF} ultimately yields a \SI{1.25}{\milli \second}, low-latency performance on on-board computers where GPUs are not available, and accuracy that rivals the performance of pose graph-based SLAM methods. 
Further, extensive in-field testing revealed that while under nominal grip levels the racing performance of a pose-graph SLAM method such as \emph{Cartographer} is superior, with low-quality odometry signal the MCL-based \emph{SynPF} shows robustness (-0.08\% scan alignment and -6.9\% lateral error) whereas the SLAM counterpart significantly worsens (-11.0\% scan alignment score and +66.6\% lateral tracking error). This allows operators to determine a priori to a race which kind of localization algorithm would be most suited for the given case. 

\bstctlcite{IEEEexample:BSTcontrol} 
\bibliographystyle{IEEEtranDOI}
\bibliography{main}



\end{document}